\PassOptionsToPackage{table,xcdraw,dvipsnames}{xcolor}
\PassOptionsToPackage{round}{natbib}
\documentclass[11pt,letterpaper]{plum}

\usepackage{natbib}

\usepackage{latexsym}
\usepackage{amsmath}
\usepackage{amssymb}
\usepackage{subcaption}
\usepackage{enumitem}
\usepackage{tabularx}
\usepackage{booktabs}
\usepackage{multirow}
\usepackage{placeins}
\usepackage{wrapfig}
\usepackage{graphicx}
\usepackage{makecell}
\usepackage{soul}
\usepackage{xcolor}
\usepackage{fontawesome5} 
\tcbuselibrary{skins,breakable,listings}
\usepackage{algorithm}
\usepackage{algpseudocode}
\definecolor{deltapos}{RGB}{20,90,28}   
\definecolor{deltaneg}{RGB}{170,20,20}   
\definecolor{promptPurple}{HTML}{5D1548} 

\newtcolorbox{promptbox}[2][]{%
    enhanced,
    breakable,     
    colback=white,      
    colframe=promptPurple, 
    coltitle=white,     
    fonttitle=\bfseries\large,
    title={\faInfoCircle\ \ #2}, 
    arc=3mm,            
    boxrule=1.5pt,      
    left=5pt, right=5pt, top=5pt, bottom=5pt,
    overlay first={
        \draw[promptPurple, line width=1.5pt] (frame.south west) -- (frame.south east); 
    },
    overlay middle={
        \draw[promptPurple, line width=1.5pt] (frame.north west) -- (frame.north east);
        \draw[promptPurple, line width=1.5pt] (frame.south west) -- (frame.south east);
    },
    overlay last={
        \draw[promptPurple, line width=1.5pt] (frame.north west) -- (frame.north east);
    },
    #1
}

\newtcblisting{promptlisting}[1]{%
    enhanced, breakable, listing only,
    colback=white, colframe=promptPurple, coltitle=white,
    fonttitle=\bfseries, title={\faInfoCircle\ \ #1},
    arc=2mm, boxrule=1pt,
    left=4pt, right=4pt, top=3pt, bottom=3pt,
    listing options={basicstyle=\ttfamily\scriptsize\color{black!85}, breaklines=true,
        breakatwhitespace=false, breakindent=0pt, columns=fullflexible, keepspaces=true,
        showstringspaces=false, upquote=true, aboveskip=0pt, belowskip=0pt,
        emph={Thought,Action}, emphstyle={\bfseries\color{promptPurple}},
        moredelim={[s][\color{caseStudent}]{<}{>}}},
}

\definecolor{caseStudent}{HTML}{2B6CB0}
\definecolor{caseTeacher}{RGB}{199,67,117}
\definecolor{caseEnv}{HTML}{5A5A5A}
\newtcolorbox{caseturn}[3]{
    enhanced, breakable=false,
    colback=#1!4, colframe=#1, boxrule=0pt, leftrule=2.5pt, arc=0pt,
    left=5pt, right=4pt, top=2pt, bottom=2pt,
    before skip=3pt, after skip=3pt,
    fontupper=\small,
    before upper={{\color{#1}\bfseries #2}\hfill{\color{#1}\footnotesize #3}\par\vspace{1pt}},
}
\newcommand{\caseT}[1]{\textit{\color{black!70}Thought:} #1\par}
\newcommand{\caseA}[1]{\textit{\color{black!70}Action:} \texttt{#1}\par}
\newcommand{\caseO}[1]{\textit{\color{black!70}Obs:} {\color{caseEnv}#1}\par}
\newcommand{\casehandoff}[1]{%
    \par\vspace{4pt}\noindent
    {\color{caseTeacher}\leaders\hrule height 0.8pt\hfill\ \textbf{\small\faExchange*\ #1}\ \leaders\hrule height 0.8pt\hfill}%
    \par\vspace{4pt}}

\newcommand{\casetrim}{{\color{black!45}[\ldots]}}
\newcommand{\casecut}{\casehandoff{Cut at student token 4096: teacher continues from the same character}}

\definecolor{hl_yellow}{HTML}{FFF3A8}
\definecolor{hl_blue}{HTML}{DAE8FC}
\definecolor{hl_pink}{HTML}{FFE3ED}
\definecolor{hl_green}{HTML}{D4F5E3}
\definecolor{rose}{RGB}{199,67,117}

\newcommand{\hlp}[1]{%
  {\sethlcolor{hl_pink}\hl{#1}}%
}
\newcommand{\hlg}[1]{%
  {\sethlcolor{hl_green}\hl{#1}}%
}

\usepackage{microtype}
\ifdefined\XeTeXversion
  \microtypesetup{tracking=false}
\fi

\usepackage{inconsolata}

\usepackage{xspace}

\usepackage{hyperref}
\usepackage{url}
\usepackage{cleveref}
\usepackage{pifont}

\crefname{section}{\S\!}{\S\S\!}
\crefname{table}{Tab.}{Tabs.}
\crefname{figure}{Fig.}{Figs.}
\crefname{algorithm}{Alg.}{Algs.}
\crefname{appendix}{App.}{Apps.}
\crefname{equation}{Eq.}{Eqs.}

\newtcolorbox{takeaway}{
  enhanced, breakable=false,
  colback=white, colframe=promptPurple,
  boxrule=0.8pt, arc=2pt,
  left=6pt, right=6pt, top=3pt, bottom=3pt,
  before skip=6pt, after skip=6pt,
  before upper={{\color{promptPurple}\bfseries Takeaway.}\ },
}

\definecolor{heat}{HTML}{6C63D9}
\newcommand{\hc}[2]{\cellcolor{heat!#1}#2}
\newcommand{\tn}[1]{{\color{black!55}#1}} 
\newcommand{\dpos}[1]{\textcolor{deltapos}{\scriptsize(+#1)}}
\newcommand{\dneg}[1]{\textcolor{deltaneg}{\scriptsize($-$#1)}}

\newcommand{\method}{\textsc{OLIVE}\xspace}

\title{Learning from Teacher Continuations at Student States\\[0.4em]
  {\normalfont\sffamily\bfseries\fontsize{11}{13}\selectfont Ongoing work}}
\author[1,*]{Haojin Wang}
\author[1,*]{Dylan Zhang {\color{MyHeaderText}\textbf{(Project lead)}}}
\author[2]{Huaibo Chen}
\author[3]{Suhao Yu}
\author[1]{Yihang Sun}
\author[1]{Zhanyang Jin}
\author[4]{Jiaying Ye}
\makeatletter
\let\ABorig@authnote\AB@authnote
\renewcommand\AB@authnote[1]{}
\author[1]{Dianqi Li}
\let\AB@authnote\ABorig@authnote
\makeatother
\author[5]{Prasanna Sattigeri}
\author[2]{Kamal Youcef-Toumi}
\author[1]{Hao Peng}
\affil[1]{University of Illinois at Urbana-Champaign}
\affil[2]{Massachusetts Institute of Technology}
\affil[3]{University of Pennsylvania}
\affil[4]{University of Washington}
\affil[5]{International Business Machines}
\makeatletter
\g@addto@macro\AB@affillist{\\[0.5em]\Affilfont
  \textsuperscript{*}Main Contributors, Equal Contribution.}
\makeatother
\paperurl{https://dylanzsz.github.io/olive/}
\definecolor{darkblue}{rgb}{0,0,0.5}
\hypersetup{colorlinks=true,citecolor=darkblue,linkcolor=darkblue,urlcolor=darkblue}
\hypersetup{pdftitle={Learning from Teacher Continuations at Student States},
  pdfauthor={Haojin Wang, Dylan Zhang, Huaibo Chen, Suhao Yu, Yihang Sun, Zhanyang Jin, Jiaying Ye, Dianqi Li, Prasanna Sattigeri, Kamal Youcef-Toumi, Hao Peng}}

\newcommand{\institutionlogos}{%
  \raisebox{-0.5\height}{\includegraphics[height=20pt]{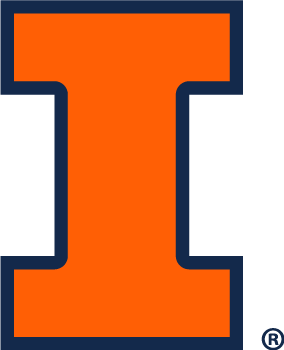}}\hspace{18pt}%
  \raisebox{-0.5\height}{\includegraphics[height=18pt]{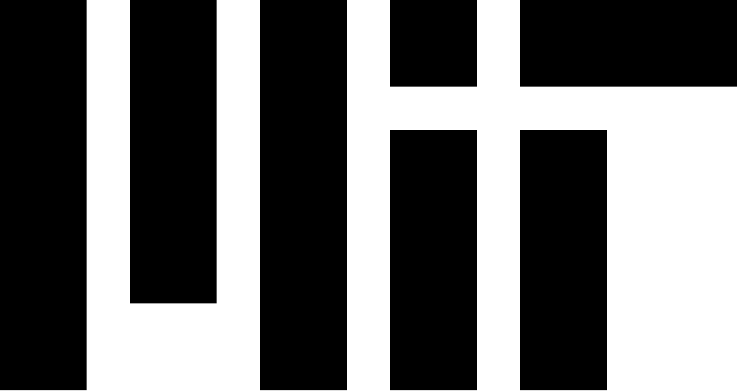}}\hspace{18pt}%
  \raisebox{-0.5\height}{\includegraphics[height=17pt]{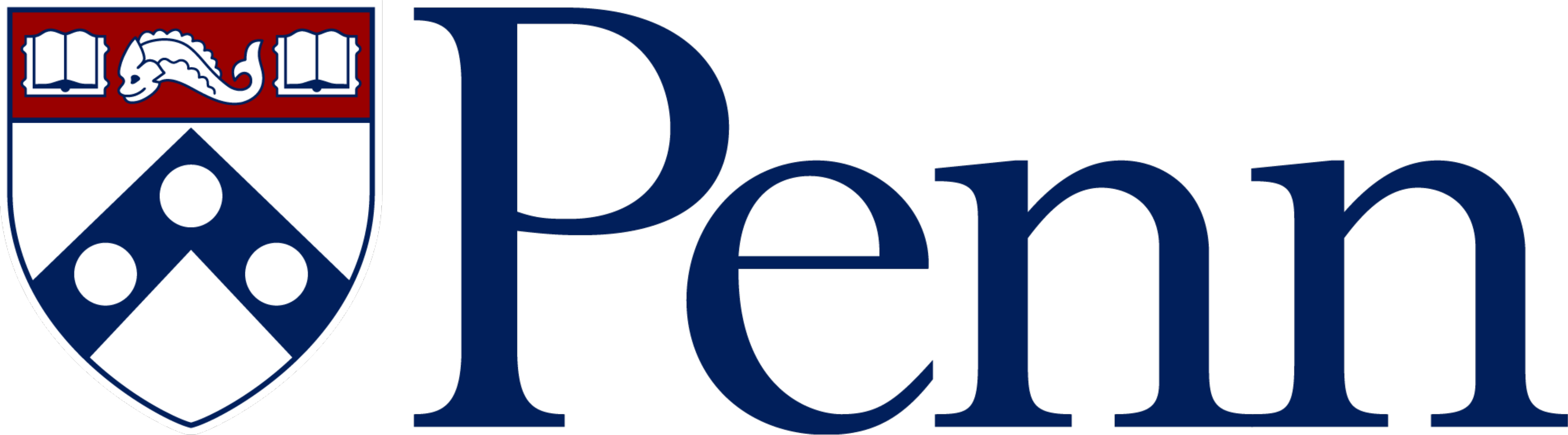}}\hspace{18pt}%
  \raisebox{-0.5\height}{\includegraphics[height=15pt]{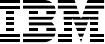}}\hspace{18pt}%
  \raisebox{-0.5\height}{\includegraphics[height=19pt]{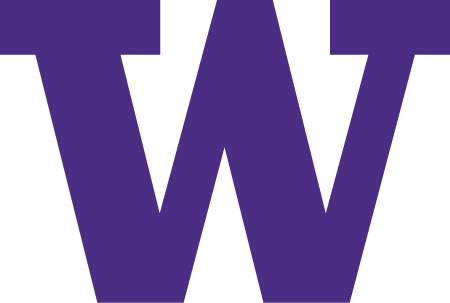}}%
}
\fancypagestyle{olivefirst}{%
  \fancyhead[L]{\institutionlogos}%
  \fancyhead[C]{}%
  \fancyhead[R]{\href{https://dylanzsz.github.io/olive/}{\urlheaderfont\itshape https://dylanzsz.github.io/olive/}}%
  \fancyfoot{}%
}

\begin{abstract}

We present \method (Online Interventional Distillation).
At each iteration, the evolving student policy generates a new prefix, the teacher continues it autoregressively, and the student is updated using cross-entropy computed on the teacher-generated tokens.
Each design choice targets a corresponding limitation of existing distillation methods: (1) sequential covariate shift in offline supervised fine-tuning (SFT) on fixed teacher trajectories, (2) fragmented supervision under prefix failure in token-level on-policy distillation (OPD), and (3) the need for access to teacher token probabilities in distribution-matching distillation.
\method achieves higher reasoning performance than OPD (with a top-16 KL approximation) at comparable GPU-hour cost.
Our asynchronous implementation further reduces \method's total training time by 23.8\%.
We evaluate \method on both hard reasoning tasks and agentic tasks which reflects modern post-training scenarios, and it consistently outperforms existing distillation methods under the same training budget.
By regenerating prefixes from the evolving student, \method continues improving after offline distillation plateaus while better preserving the general capabilities and plasticity of the student.
Using only text from GPT-5.4-mini, continuously training with \method outperforms offline SFT from the same teacher by 13\% on ScienceWorld.
These results support \method as an effective and efficient approach to online language-model distillation.

\end{abstract}

\begin{document}
\maketitle
\thispagestyle{olivefirst}


\section{Introduction}
\label{sec:intro}

Knowledge distillation transfers capability from a stronger teacher to a weaker student, either by matching the teacher’s token-level distributions or by training on text the teacher generates~\citep{hinton2015distilling,kim2016sequence,west2022symbolic}.
Offline supervised fine-tuning (SFT) trains the student on fixed teacher trajectories, whereas at inference it conditions on its own outputs.
This sequential covariate shift can cause errors to compound over long horizons~\citep{pmlr-v9-ross10a,bengio2015scheduled}.
As the student policy changes during training, a fixed dataset also fails to track the states it currently visits.
Offline SFT can also degrade the student's prior capabilities~\citep{shenfeld2026rl,chen2025retaining}.
Prior work suggests that supervision close to the student's own distribution can improve adaptation and reduce forgetting~\citep{zhang2026best,chen2025retaining}.

Recently, on-policy distillation (OPD) has become a promising paradigm for large language model (LLM) post-training~\citep{agarwal2024policy,yang2025qwen3,lu2025onpolicydistillation,xiao2026mimo}.
By sampling rollouts from the student policy itself, OPD uses the teacher policy to calculate the reverse-KL loss for each token in the rollout.
It thus pairs dense supervision with on-policy states, anchoring learning where the student actually is rather than pulling it toward teacher trajectories~\citep{lu2025onpolicydistillation}.
Yet token-level OPD computes teacher targets along each sampled student rollout without revising it.
Even when the teacher recommends changing a token, subsequent targets remain conditioned on the student's original continuation.
The supervision therefore does not directly demonstrate how to continue from that correction~\citep{jiang2026trajectory}.
Recent works have also shown that as OPD transfers supervision from teacher at distribution level, it assumes the teacher places meaningful probability mass on the states the student reaches~\citep{zhu2026many,opd2026coldstart}.
Such assumption fails once the capability gap between the two policies is too large, and it extends to multi-turn agentic tasks where the irreversible actions made by the student lead the whole trajectory out of the support from the teacher~\citep{wang2026tcod}.
Distribution-matching OPD also requires access to teacher token probabilities, limiting its use with teachers that expose only generated text.

\begin{figure}[t]
    \centering
    \includegraphics[width=0.95\linewidth]{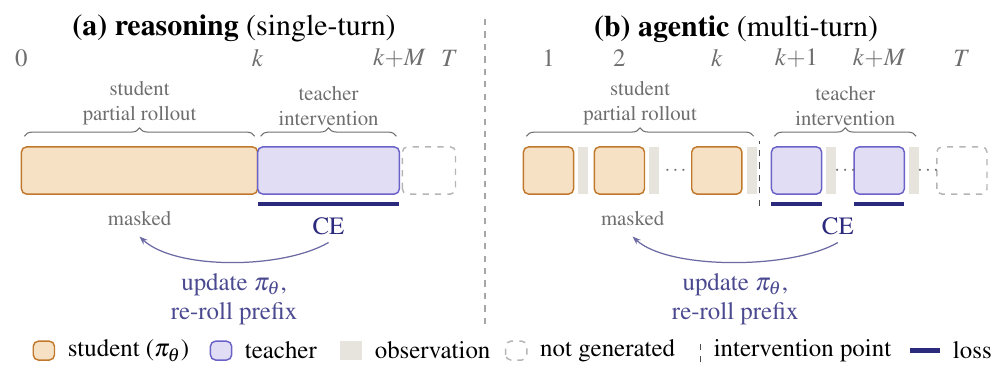}
    \caption{Overview of \method.
    \textbf{(a) Long chain-of-thought reasoning tasks:} the student generates a reasoning prefix, which the teacher continues.
    \textbf{(b) Long-horizon agentic tasks:} the student interacts with the environment for part of an episode, then the teacher takes over.
    In both settings, prefixes are refreshed as the student policy changes.
    }
    \vspace{-10pt}
    \label{fig:teaser}
\end{figure}

We propose \method (Online Interventional Distillation).
It trains the evolving student on teacher continuations from student-generated states (\cref{fig:teaser}).
At each iteration, the evolving student policy generates a new prefix, the teacher continues it autoregressively, and the student is updated using cross-entropy computed only on the teacher-generated tokens. For agentic tasks, the prefix consists of the student's actions and the resulting observations; the teacher then takes over interaction with the environment.
Repeating this process refreshes the prefixes as the student improves.
To reduce the time spent waiting for teacher generation, we implement \method asynchronously, drawing on asynchronous reinforcement learning~\citep{mnih2016asynchronous,espeholt2018impala}.
Teacher continuation for one batch overlaps with student prefix generation for the next (\cref{fig:async}), reducing total training time by 23.8\% relative to synchronous \method (\cref{sec:efficiency}).

We evaluate \method in the two settings at the center of frontier post-training~\citep{team2026kimi, xu2026deepseek, zeng2026glm}: long chain-of-thought reasoning and long-horizon agentic tasks.
For reasoning, we target problems beyond the student's capability, using synthetic tasks from RLVE~\citep{zeng2025rlve} whose difficulty we can control.
For agentic tasks, we use multiple environments from AgentGym~\citep{xi2025agentgym}.
Empirically, we find that \method lifts the performance of the student policy on hard reasoning tasks by 6\% to 8\% pass@8 points and 7\% to 22\% avg@4 gains on agentic benchmarks, while only introducing 0.9\% average performance drop on general benchmarks (\cref{sec:experiment}, \cref{sec:online_selection}).
The teacher continuation enables \method to learn where OPD would fail, raising ScienceWorld success to 7.5\% from a near-zero student that OPD fails to improve (\cref{sec:agentic}).
With the nature of CE loss, \method does not require the teacher logits, enabling black-box distillation (\cref{sec:moving_policy}).
Refreshing the student policy online, we show that \method enables progressive performance improvement as the training proceeds, while offline distillation plateaus and fails to maintain plasticity (\cref{sec:moving_policy}).

Taken together, we present \method, an online distillation method that trains the rolling student policy on teacher continuations elicited at the states its own prefixes reach.
\method moves the optimization objective in SFT from offline to online, so the supervision is refreshed as the student improves.
On top of OPD, \method demonstrates what to do next from student-visited states, rather than grading past tokens, and needs only teacher text.
Empirically, \method lets training keep improving where offline distillation plateaus, offering better plastcity while causing less forgetting of the capability it already has.
Our results suggest that where supervision is placed, and whether it is refreshed as the student changes, is an important axis of distillation design alongside the form the supervision takes.

\section{Motivation}
\label{sec:background}
Distillation provides supervision through teacher distributions or generated texts~\citep{hinton2015distilling,kim2016sequence,west2022symbolic}, but its usefulness also depends on the contexts at which that supervision is provided.
For difficult reasoning and interaction tasks, we ask: \textbf{how should supervision from the teacher connect the student's own attempts to behavior it does not yet generate reliably?}

\paragraph{Teacher continuations at student-generated contexts.}
Training only on teacher trajectories can leave the student unprepared for situations created by its own decisions, a source of compounding errors in behavioral cloning~\citep{pomerleau1991efficient,ross2011reduction}.
OPD addresses this by supervising the student along its own rollouts~\citep{agarwal2024policy}, but later targets remain conditioned on the student's earlier decisions even where the teacher recommends a different one, so the rollout never demonstrates what would follow that recommendation~\citep{jiang2026trajectory}.
On-policy supervision also struggles under large capability gaps and when student errors derail multi-turn interaction~\citep{opd2026coldstart,zhu2026many,wang2026tcod}.
Teacher takeover instead lets the teacher continue from a student-generated prefix, so later decisions, and in interactive environments later observations, follow the teacher's own choices.
The student thus learns how to proceed from contexts it actually reaches, without first having to produce the corrective path itself.
This mirrors learner roll-in with expert rollout in imitation learning~\citep{ross2014reinforcement} and recent expert-intervention methods for language-model agents~\citep{lauffer2025imitation,li2026revisiting}.

\paragraph{Online intervention with a rolling policy.}
Prefixes collected once reflect the behavior of an earlier student.
As training proceeds to update the student policy, the student may encounter different situations and benefit from different continuations.
We therefore regenerate prefixes and teacher interventions throughout training, following the learner-state supervision principle of DAgger~\citep{ross2011reduction}.
\citet{zhang2026best} and \citet{zhang2026good} likewise motivate assessing supervision in relation to the target student and its subsequent learning.
We examine whether refreshing intervention contexts improves learning over repeatedly using interventions collected from the initial policy.

\paragraph{Learning through teacher-generated text.}

Teacher intervention comes in textual forms, learning it only requires cross-entropy loss.
The teacher need not expose token probabilities, and its output can be tokenized using the student's tokenizer.
Teacher continuation determines the trajectory to learn from; suffix CE makes learning from that trajectory possible through a text-only interface.

Together, these choices motivate \method as a continuously refreshed procedure for learning from teacher continuations of student attempts.
We evaluate its effectiveness on complex reasoning and long-horizon interaction tasks, together with the cost of generating these interventions.

\section{OLIVE: Online Interventional Distillation}
We describe \method in \cref{sec:main} and extend it to multi-turn agentic tasks in \cref{sec:multi-turn}.
We then present an asynchronous implementation that overlaps student sampling with teacher generation to reduce idle time (\cref{sec:async}).

\subsection{Main Algorithm}\label{sec:main}
\begin{wrapfigure}[11]{R}{0.5\textwidth}
  \vspace{-2.6em}
  \begin{minipage}{0.5\textwidth}
  \begin{algorithm}[H]
    \caption{\method (Online Interventional Distillation)}
    \label{alg:sod}
    \begin{algorithmic}[1]
    \footnotesize
    \Require Student $\pi_\theta$, teacher $\pi_T$, prompts $\mathcal{D}$, student prefix length $k$, teacher continuation budget $M$.
    \For{each training step}
        \State Sample prompts $x \sim \mathcal{D}$
        \State \textbf{Online Roll-in:} $y_{1:k} \sim \pi_\theta(\cdot \mid x)$
        \State \textbf{Continue:} $\tilde{y} \sim \pi_T(\cdot \mid x, y_{1:k})$, $|\tilde{y}| \le M$
        \State \textbf{Update:} Update $\theta$ using the masked CE loss in \cref{eq:osd-ce}.
    \EndFor
    \State \Return $\pi_\theta$
    \end{algorithmic}
    \vspace{0.1em}
  \end{algorithm}
  \end{minipage}
  \vspace{-50pt}
\end{wrapfigure}
\method trains a student $\pi_\theta$ from a teacher $\pi_T$ by collecting student prefixes online, letting the teacher demonstrate how to continue, and learning from teacher text alone.
These three design choices address the limitations of offline SFT and token-level OPD, as we explain below using \cref{alg:sod} as a walkthrough.

\vspace{-6pt}
\paragraph{Collect student prefixes online.}
To obtain supervision at states reached by the current student, we sample a prompt $x$ from the prompt distribution $\mathcal{D}$ and a $k$-token prefix $y_{1:k} \sim \pi_\theta(\cdot \mid x)$ (\cref{alg:sod}, lines 2--3).
We repeat this step after each student update, so the supervision tracks the evolving policy instead of remaining tied to fixed offline trajectories.
We use a fixed prefix length $k$ and retain all sampled prefixes without filtering.
We report the student and teacher generation lengths for reasoning and agentic tasks in \cref{sec:reasoning,sec:agentic}, respectively.

\paragraph{Demonstrate how to continue.}
Token-level OPD leaves later targets conditioned on the student's original tokens even when an earlier target recommends a correction.
In line 4 of \cref{alg:sod}, the teacher receives the prompt and student prefix as the input and generates a continuation $\tilde{y} \sim \pi_T(\cdot \mid x,y_{1:k})$ of at most $M$ tokens under its own tokenizer.
The teacher conditions each new token on its preceding choices, allowing the continuation to demonstrate how to follow a correction when the student prefix remains recoverable.
We use partial continuations to bound the cost of online teacher generation, without requiring a completed, verified solution.
As we show in \cref{sec:efficiency}, \method achieves higher reasoning performance than OPD at comparable GPU-hour cost with this limited continuation budget.

\paragraph{Learn from teacher text alone.}
To avoid requiring teacher token probabilities, we train the student with CE on the generated continuation (\cref{alg:sod}, line 5).
For the student update, we tokenize the teacher continuation with the student's tokenizer, obtaining $\tilde{y}_{1:\ell}$, where $\ell$ is its length in student tokens. This length may differ from the number of teacher tokens. We feed the full sequence $(x,y_{1:k},\tilde{y}_{1:\ell})$ to the student and compute
\begin{equation}
    \mathcal{L}(\theta; x,y_{1:k},\tilde{y}_{1:\ell})
    = -\sum_{j=1}^{\ell}
    \log \pi_\theta\!\left(\tilde{y}_j \mid x,y_{1:k},\tilde{y}_{<j}\right).
    \label{eq:osd-ce}
\end{equation}
The prompt and student prefix remain in the conditioning context but are masked out of the loss; the sampled sequences are held fixed during the update.
This objective requires only teacher-generated text, so it supports black-box teachers and different teacher and student tokenizers.

\begin{figure}[t]
    \centering
    \includegraphics[width=0.9\linewidth]{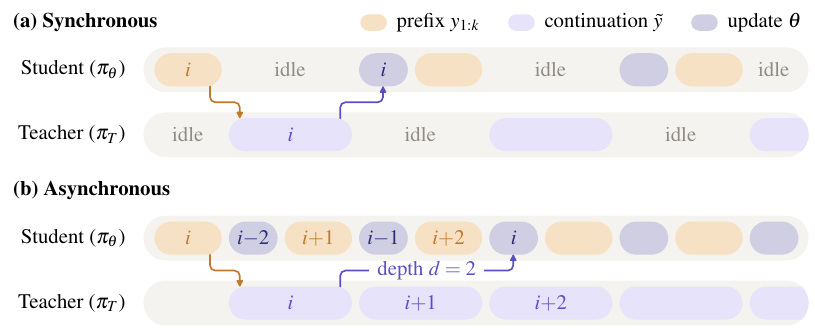}
    \caption{Synchronous and asynchronous implementations of \method. (a) The student waits for the teacher continuation before updating on the same batch. (b) Prefix sampling for later batches overlaps with teacher generation, while student updates use completed continuations from earlier batches. Arrows connect teacher continuations to the updates that use them. }
    \label{fig:async}
\end{figure}

\subsection{Multi-turn \method}
\label{sec:multi-turn}
\method extends to multi-turn interaction by applying the same prefix--continuation split at the level of turns (\cref{fig:teaser}b). Here, $k$ and $M$ count interaction turns rather than tokens.
The student interacts with the environment for $k$ turns, producing a history of actions and observations.
The teacher then takes over for up to $M$ turns, with each action conditioned on the full interaction history and executed in the environment to obtain the next observation.
During the student update, we retain the full interaction history as context, mask the student's turns and all environment observations, and apply CE only to the teacher's actions.
We find that student prefixes of $k=5$ or $10$ turns, depending on the environment, followed by up to $M=5$ teacher turns work well (\cref{sec:agentic}).

\subsection{Asynchronous \method for Scalable Training}
\label{sec:async}
To reduce student GPU idle time, we overlap student prefix sampling with teacher generation (\cref{fig:async}), drawing on asynchronous reinforcement learning~\citep{mnih2016asynchronous,espeholt2018impala}.
While the teacher generates continuations for one batch, the student samples prefixes for the next; completed traces provide training data for the same masked CE objective in \cref{eq:osd-ce}.
This overlap means that a prefix may come from an earlier student policy than the one being updated.
We bound this lag by an asynchronous depth $d$, the maximum number of student updates between prefix generation and use of the resulting trace for training.
A larger $d$ allows more overlap but permits greater mismatch between the policy that generated the prefix and the policy being trained.
We use $d=3$ in the reasoning experiments (\cref{tab:rose-config}) and evaluate the efficiency--performance tradeoff in \cref{sec:efficiency}.

\section{Experiments}
\label{sec:experiment}
We evaluate \method on two main post-training scenarios.
We present the experiment in reasoning tasks in \cref{sec:reasoning} and multi-turn agentic tasks in \cref{sec:agentic}.
We further study the training efficiency of \method in \cref{sec:efficiency}.

\subsection{Reasoning Task}
\label{sec:reasoning}
\paragraph{Task.}
We use RLVE~\citep{zeng2025rlve} as our primary testbed for single-turn reasoning tasks.
RLVE is a synthetic reasoning environment that consists of different reasoning environments, each with verifiable rewards and different difficulty parameters to construct problems with tunable difficulty.
It provides a noise-free data collection, training and evaluation pipeline as the instances are not provided during pre-training or post-training of the models themselves to introduce contamination~\citep{shao2025spurious}.
Since knowledge distillation targets at introducing new capabilities from the teacher policy to the student policy, we set the difficulty parameters to find the problems that are challenging enough for the student policy to solve.
We thus obtain a 18 different reasoning environments subset with 500 training problems for each game, yielding a pool of $9$K hard problems.
For each task, we pair with 10 test problems with the same difficulty as the training problems.

\begin{table}[t]
    \centering
    \begin{minipage}[b]{0.585\textwidth}
    \centering
    \small
    \setlength{\tabcolsep}{3pt}
    \renewcommand{\arraystretch}{1.1}
    \begin{tabular}{lcccc}
    \toprule
     & \multicolumn{2}{c}{\textbf{Qwen3-1.7B}} & \multicolumn{2}{c}{\textbf{Qwen3-4B}} \\
    \cmidrule(lr){2-3} \cmidrule(lr){4-5}
    \textbf{Method} & Pass@8 & Avg@8 & Pass@8 & Avg@8 \\
    \midrule
    Original & \hc{4}{11.1} & \hc{4}{3.3} & \hc{6}{46.1} & \hc{4}{18.1} \\
    \midrule
    Teacher-Gen. & \hc{21}{15.0 \dpos{3.9}} & \hc{22}{5.6 \dpos{2.3}} & \hc{40}{\textbf{53.3} \dpos{7.2}} & \hc{16}{19.9 \dpos{1.8}} \\
    OPD & \hc{18}{14.4 \dpos{3.3}} & \hc{14}{4.6 \dpos{1.3}} & \hc{4}{45.6 \dneg{0.5}} & \hc{25}{21.3 \dpos{3.2}} \\
    \midrule
    \textbf{\method} & \hc{40}{\textbf{19.4} \dpos{8.3}} & \hc{40}{\textbf{7.8} \dpos{4.5}} & \hc{35}{52.2 \dpos{6.1}} & \hc{40}{\textbf{23.5} \dpos{5.4}} \\
    \bottomrule
    \end{tabular}
    \caption{Results on RLVE. We report pass@8 and avg@8 on the test set, with two different student models thinking enabled. We use identical teacher model Qwen3-4B-Thinking-2507 for all distillation methods. \method uses the asynchronous implementation with $d{=}3$. All numbers are percentages ($\%$).}
    \label{tab:rlve}
    \end{minipage}\hfill
    \begin{minipage}[b]{0.4\textwidth}
    \centering
    \includegraphics[width=\linewidth]{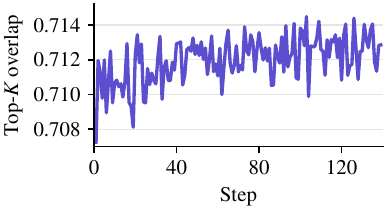}
    \captionof{figure}{Top-$K$ overlap ratio between student and teacher on validation set during OPD training (Qwen3-1.7B student, Qwen3-4B-Thinking-2507 teacher). It stays nearly flat ($0.707\to0.713$), resonating the finding in \citet{opd2026coldstart}. }
    \label{fig:opd_overlap}
    \end{minipage}
\end{table}

\paragraph{Models and Evaluation.}
We consider using models with thinking capabilities that are able to solve the reasoning problems with long-term reasoning generation.
To achive this, we use Qwen3-1.7B and Qwen3-4B~\citep{yang2025qwen3} as two student models with their thinking enabled.
We use Qwen3-4B-Thinking-2507~\citep{yang2025qwen3} as the teacher policy since it is the continued scaled model from Qwen3-4B and has stronger reasoning capabilities.
We report the pass@8 and avg@8 on the test set.

\paragraph{Training Setup.}
We report the results of comparison between \method and other distillation methods, including offline distillation and OPD in \cref{tab:rlve}.
To compare distillation methods under a matched budget, we cap the total number of distilled tokens per rollout at 7168 for both the offline and online baselines.
The offline baseline applies cross-entropy on unfiltered teacher trajectories, and the online baseline is OPD.
Since \method does not require the student policy to generate the entire rollout, we set the prefix length to 4096 and the continuation length of 1024 by default.
We set the asynchronous depth $d = 3$ for \method.
Detailed experimental settings are provided in the \cref{app:impl}.

\paragraph{Results.}
We report the results of \method and comparison across two different student models in \cref{tab:rlve}.
Across both students, \method achieves the largest avg@8 gains among all distillation methods (+4.42 and +5.35 points).
Compare to offline distillation where offline data is collected from the teacher policy without filtering, \method achieves better improvements with less samling from the teacher policy and remain non-filtering.
We also include a training dynamic visualiztion of top-$K$ overlap ratio between student and teacher on validation set during OPD training in \cref{fig:opd_overlap}.
Resonating the finding in \citet{opd2026coldstart}, the top-$K$ overlap ratio remains nearly flat during OPD training, and little improvement is observed in OPD during training due to different thinking behaviors between the student and the teacher.
Overall, supervising the student with teacher continuations from its own states transfers more capability than either training on full teacher trajectories or scoring the student's rollouts, while requiring only a fraction of the teacher's generation.

\subsection{Multi-turn Agentic Task}
\label{sec:agentic}
\paragraph{Task.}
We select 5 multi-turn agentic tasks from AgentGym~\citep{xi2025agentgymenv}.
AgentGym is a multi-turn agentic environment that provides a diverse environments with turn-level feedback for each action the agent takes.
Following \citet{xi2025agentgym}, we select 5 environments, including ALFWorld~\citep{shridhar2020alfworld}, ScienceWorld~\citep{wang2022scienceworld}, SearchQA~\citep{dunn2017searchqa}, TextCraft~\citep{prasad-etal-2024-adapt} and BabyAI~\citep{chevalier2018babyai}.
We use the same training and evaluation set for different tasks, except for SearchQA we construct a 6K problems training set with 400 held out problems for evaluation.

\begin{table}[t]
    \centering
    \small
    \setlength{\tabcolsep}{3.5pt}
    \renewcommand{\arraystretch}{1.1}
    \begin{tabular}{l cc cc cc cc cc}
    \toprule
    & \multicolumn{2}{c}{\textbf{ALFWorld}}
    & \multicolumn{2}{c}{\textbf{ScienceWorld}}
    & \multicolumn{2}{c}{\textbf{TextCraft}}
    & \multicolumn{2}{c}{\textbf{BabyAI}}
    & \multicolumn{2}{c}{\textbf{SearchQA}} \\
    \cmidrule(lr){2-3} \cmidrule(lr){4-5} \cmidrule(lr){6-7} \cmidrule(lr){8-9} \cmidrule(lr){10-11}
    \textbf{Method} & SR ($\%$) & \tn{Turns} & SR ($\%$) & \tn{Turns} & SR ($\%$) & \tn{Turns} & SR ($\%$) & \tn{Turns} & SR ($\%$) & \tn{Turns} \\
    \midrule
    Student & \hc{4}{19.38} & \hc{3}{\tn{26.97}} & \hc{5}{0.12} & \hc{6}{\tn{28.63}} & \hc{4}{23.00} & \hc{3}{\tn{23.89}} & \hc{4}{38.33} & \hc{3}{\tn{14.52}} & \hc{8}{30.50} & \hc{7}{\tn{11.91}} \\
    {\color{black!55}\textit{Teacher}} & {\color{black!55}52.12} & {\color{black!55}21.24} & {\color{black!55}15.62} & {\color{black!55}23.61} & {\color{black!55}85.50} & {\color{black!55}10.96} & {\color{black!55}83.33} & {\color{black!55}6.22} & {\color{black!55}55.06} & {\color{black!55}9.29} \\
    \midrule
    OPD & \hc{9}{22.25} & \hc{7}{\tn{26.28}} & \hc{4}{0.00} & \hc{6}{\tn{28.77}} & \hc{11}{29.50} & \hc{8}{\tn{22.28}} & \hc{10}{43.06} & \hc{4}{\tn{14.32}} & \hc{4}{29.56} & \hc{3}{\tn{12.07}} \\
    TCoD-B2F & \hc{35}{37.10} & \hc{24}{\tn{23.30}} & \hc{8}{0.75} & \hc{4}{\tn{29.28}} & \hc{22}{39.50} & \hc{15}{\tn{19.98}} & \hc{38}{66.30} & \hc{22}{\tn{9.03}} & \hc{35}{37.69} & \hc{25}{\tn{11.12}} \\
    TCoD-F2B & \hc{19}{28.00} & \hc{15}{\tn{24.89}} & \hc{6}{0.50} & \hc{3}{\tn{29.55}} & \hc{29}{45.50} & \hc{19}{\tn{18.46}} & \hc{34}{62.50} & \hc{17}{\tn{10.51}} & \hc{35}{37.75} & \hc{23}{\tn{11.19}} \\
    Guided OPD & \hc{18}{27.12} & \hc{13}{\tn{25.26}} & \hc{8}{0.75} & \hc{5}{\tn{29.03}} & \hc{29}{45.50} & \hc{19}{\tn{18.59}} & \hc{40}{\textbf{67.78}} & \hc{25}{\tn{\textbf{8.30}}} & \hc{34}{37.50} & \hc{25}{\tn{\textbf{11.10}}} \\
    \midrule
    \textbf{\method} & \hc{40}{\textbf{40.00}} & \hc{25}{\tn{\textbf{23.15}}} & \hc{40}{\textbf{7.50}} & \hc{25}{\tn{\textbf{23.64}}} & \hc{40}{\textbf{55.25}} & \hc{25}{\tn{\textbf{16.45}}} & \hc{40}{67.50} & \hc{19}{\tn{9.98}} & \hc{40}{\textbf{39.06}} & \hc{24}{\tn{11.13}} \\
    \bottomrule
    \end{tabular}
    \caption{Results on multi-turn agentic benchmarks (ALFWorld, ScienceWorld, TextCraft, BabyAI and SearchQA) with Qwen3-1.7B as the student and Qwen3-32B as the teacher.
    We report avg@4 success rate (SR, $\%$) and average trajectory score, along with the average turns across the five benchmarks.}
    \label{tab:agentic}
\end{table}

\paragraph{Training Setup.}
\label{sec:results}
We use Qwen3-1.7B as the student model and Qwen3-32B as the teacher model.
We compare different online distillation methods on multi-turn agentic tasks, including OPD and subsequent variants that tries to adapt OPD to the multi-turn agentic setting, including two variants of TCoD~\citep{wang2026tcod} and Guided OPD~\citep{li2026policy}.
We set the maximum number of turns for ALFWorld, TextCraft and ScienceWorld to 30, 20 for BabyAI and 16 for SearchQA.
For each turn we follow the ReAct~\citep{yao2022react} framework to generate the action as AgentGym originally implemented.
We set the training epoch for ALFWorld, ScienceWorld and SearchQA to 1, and 3 for BabyAI and TextCraft respectively, since BabyAI and TextCraft have less training data to train the model.
For evaluation, we report the average@4 success rate (SR, $\%$), along with the average turns across the five benchmarks.
Empirically, we set 10 student turns as prefix for \method for environments with longer turns like ALFWorld and ScienceWorld, and 5 student turns as prefix for environments with shorter turns.
The teacher turns continuation is set to 5 for \method by default.

\paragraph{Results.}
We report the results of comparison between \method and other online distillation methods in \cref{tab:agentic}.
Across different environments, \method consistently outperforms the other two distillation methods, with each environment showing less turns to achieve higher success rate.
Empirically, we observe the biggest performance gain on ScienceWorld, which is the most complex environment among the five according to initial student policy performance.
This also explains why OPD does not work well on this environment, as the student action at early turns are likely flawed, making the subsequent turns within the same episode to be incorrect and fail the task.
Instead, \method effectively distill the teacher capabilities to the student policy by directly introducing teacher intervention at given student states, thus correcting the student episode on the right track.

\subsection{Training Efficiency}
\label{sec:efficiency}
In this section, we study the training efficiency of \method.
We conduct a training comparison between OPD, synchronous \method and asynchronous \method on a single turn reasoning task.
To be specific, following the setting in ~\cref{tab:rlve}, we use the same teacher model Qwen3-4B-Thinking-2507 for all distillation methods, and set student model to be Qwen3-4B.
We run training on the $9$K training set of RLVE on 8 H200 GPUs, and report the total GPU hours for each method.

\begin{wrapfigure}{r}{0.4\textwidth}
    \centering
    \includegraphics[width=0.4\textwidth]{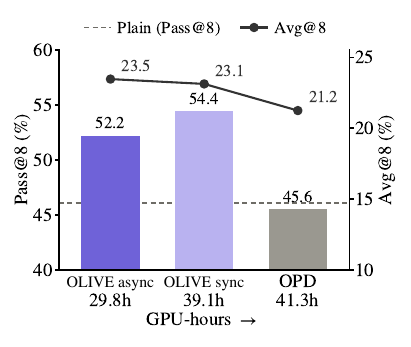}
    \caption{Training comparison between OPD, asynchronous \method and synchronous \method.}
    \label{fig:efficiency}
    \vspace{-10pt}
\end{wrapfigure}

We present the results in \cref{fig:efficiency}, and observe that \method can match the training efficiency of OPD while introducing performance improvement.
Since \method does not require the teacher model to complete the reasoning process or to compute reverse KL on all student generated tokens, it introduces less computation overhead for student rollouts and teacher continuation.
We further improves the training efficiency of \method by using asynchronous training.
Asynchronous \method lets the student policy keep sampling the next batch of prefix rollouts in parallel, and the collected traces are used for the update once the staleness limit is reached.
In practice, this significantly reduces the total training time by 28\% comparing with OPD.
Compared with synchronous \method, asynchronous \method only introduces minimal performance degradation while reducing the total training time by 23.8\%, which mitigates the effect of additional computation overhead introduced by hosting the teacher model online for sampling.

\section{Analysis}\label{sec:analysis}
\subsection{Online Prefixes Learn More while Forgetting Less}\label{sec:online_selection}
OEC~\citep{lauffer2025imitation} offers an offline variant of \method, where prefix reasoning content is sampled from the student policy, and then teacher continuation is sampled from the teacher policy.
Then they apply a filter to select only the correct reasoning content for finetuning with CE loss and student generation masked.
This also resonates SFT baseline, where the teacher model generates full reasoning content and then finetune the student policy on the generated continuations.
In this section, we present an analysis study to advocate that \textbf{online prefixes enable more effective distillation while forgetting less on general capabilities}, mitigating the exposure bias of offline distillation we mentioned in \cref{sec:intro}.

We first run teacher 4 times on the same prompts in RLVE, and filter by verifier to get the correct reasoning SFT data for finetuning.
We then collect 1 prefix rollout from the student policy for each problem in the training set, and then ask the teacher policy to carry on reasoning as continuation for 4 times, and then filter by verifier, yileding a continuation dataset for finetuning.
We finally compare this two baselines with \method, where instead of only using partial teacher continuation as in \cref{tab:rlve}, we let the teacher to generate full reasoning content and then filter during online rollouts.
The online rollout number for each problem is set to be 4, and we only compute CE loss on the correct reasoning content after filtering.

\begin{figure}[t]
    \centering
    \begin{minipage}[t]{0.48\textwidth}
        \centering
        \includegraphics[width=\linewidth]{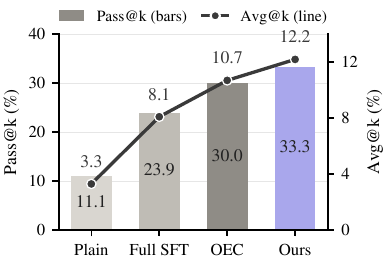}
        \caption{Comparison of \method with OEC and SFT baselines on RLVE. Obtaining prefix reasoning content from online rolling policy instead of collecting offline prefix or full reasoning content creates effective distillation.}
        \label{fig:oec}
    \end{minipage}
    \hfill
    \begin{minipage}[t]{0.48\textwidth}
        \centering
        \includegraphics[width=\linewidth]{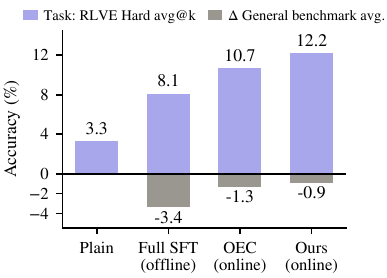}
        \caption{Evaluation of forgetting and task performance gains.
        Purple bars show pass@8 on the RLVE test set after finetuning.
        Grey bars show the change in average accuracy on general benchmarks relative to the plain student. }
        \label{fig:forgetting}
    \end{minipage}
\end{figure}

We report the results in \cref{fig:oec}.
We observe that \method achieves a better performance than both baselines, indicating the effectiveness of applying online intervention during training.
Instead of collecting the prefix reasoning content directly from the student policy as in OEC, \method collects the prefix reasoning from the updating student policy during online stage.
This effectively boost the performance of distillation by providing more effective supervision during online training.
We also measure the forgetting effect of \method on RLVE.
To directly measure the effect of such forgetting between online and offline variants, we test the performance of the finetuned policies on 4 different general benchmarks, including AIME25~\citep{balunovic2026matharena} for math tasks, LiveCoding Bench v6~\citep{jain2025livecodebench} for code tasks, IF-Eval~\citep{zhou2023instruction} for instruction following tasks and GPQA Diamond~\citep{rein2023gpqa} for science tasks.
We use avg@16 for math tasks, avg@8 for code tasks, and report the average performance change across the four benchmarks.
The results are shown in \cref{fig:forgetting}.
We observe that \method achieves less forgetting on general capabilities while maintaining higher task performance, indicating the advantage of online distillation over offline distillation.

\begin{takeaway}
Online prefixes enable more effective distillation than its offline variants, and mitigate the forgetting on general capabilities compared to offline distillation.
\end{takeaway}

\begin{figure}[H]
    \centering

    \begin{subfigure}{0.48\textwidth}
        \centering
        \includegraphics[width=\linewidth]{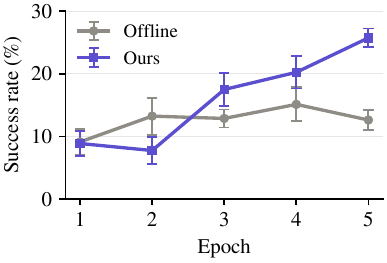}
        \caption{Over training epochs.}
        \label{fig:moving_epoch}
    \end{subfigure}
    \hfill
    \begin{subfigure}{0.48\textwidth}
        \centering
        \includegraphics[width=\linewidth]{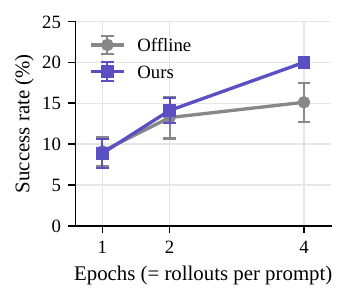}
        \caption{Over rollouts per prompt.}
        \label{fig:moving_rollout}
    \end{subfigure}
    \caption{Success rate on ScienceWorld over training 5 epochs and comparison with rollout number, with Qwen3-1.7B as the student and GPT-5.4-mini as the teacher. }
    \label{fig:moving_policy}

\end{figure}

\subsection{Online Interventions Keep the Student Learning}
\label{sec:moving_policy}
In this section we discuss the advantage of optimization under moving policy.
With only sampled text from the teacher, suffix CE turns an API model into an online teacher.
We use Qwen3-1.7B as the student policy, and GPT 5.4-mini~\citep{openai2026gpt54mini} as the teacher policy.
OPD and logit-based distillation are inapplicable here, leaving offline distillation from the same teacher as the baseline.
We use ScienceWorld as the primary benchmark for study this problem.
We first collect one episode for each problem in the training set from the teacher policy to get a training set for offline distillation.
We then use the same configuration as in \cref{sec:agentic} to train the student policy using \method.
Offline distillation is conducted on static offline data, while \method collects data as the prefix turns sampling from dynamic student policy.
We train the student policy for 5 epochs for both methods, and report the performance as the training progress.

The results are shown in \cref{fig:moving_policy}.
Training on static offline data leads to a plateau after 2 epochs, as the static data is not updated to follow the moving student policy.
Unlike offline distillation where the supervision elicitation context comes solely from static offline prompts, \method collects supervision from rolling student policy.
Such rolling policy naturally introduces diverse and suitable supervision for the student policy~\citep{zhang2026best}.
Empirically, although \method does not perform as well as offline distillation at first two epochs, it keeps improving through the training process, and successfully surpasses the performance of offline distillation by 13\% after 5 epochs.
\method is also more flexible in single epoch training.
Here we compare the performance of \method using multi-rollout and compare with corresponding offline epoch checkpoint.
We hypothesize that adding more rollouts per prompt creates same update steps as the offline distillation, yet since directly sampling from the student policy, \method can collect more diverse and suitable supervision for the student policy.
We observe that \method with multi-rollout performs better than the offline checkpoints, indicating the advantage of distillation during online stage over offline distillation.

\begin{wrapfigure}{r}{0.38\textwidth}
    \centering
    \vspace{-10pt}
    \includegraphics[width=0.38\textwidth]{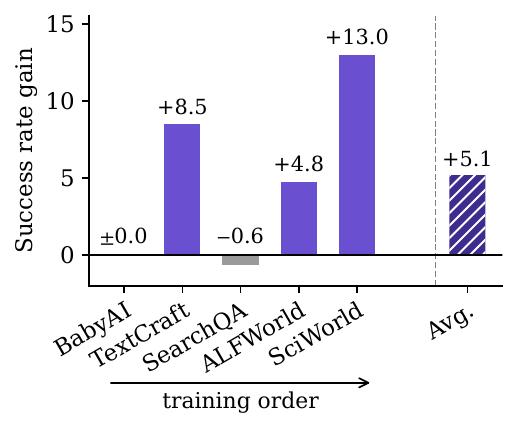}
    \caption{Success rate after 5 epochs of sequential training on each agentic environment. }
    \vspace{-15pt}
    \label{fig:plasticity}
\end{wrapfigure}

We also show \method preserves plasticity of the student policy than offline distillation methods.
For the agentic environments we used in \cref{sec:agentic}, we run offline distillation and \method for 5 epochs with the same configuration in \cref{sec:agentic} sequentially on Qwen3-4B.
After this sequential training, we evaluate the performance of the finetuned policy on the test set of each environment, obtaining \cref{fig:plasticity}.
Sequentially training on offline data leaves the student less able to learn later environments: the gap is largest on SciWorld, the last environment in the sequence, where \method improves over offline distillation by 13.0 points.
Because later environments are learned by a policy already shifted by earlier stages, fixed offline trajectories increasingly mismatch the states that policy visits, whereas \method elicits supervision from the current policy's own prefixes.
This suggests that maintaining \method online benefits the student policy to learn more effectively as different tasks proceed, demonstrating the advantage for plasticity preservation for student policy.

\begin{takeaway}
The online interventions in \method keep the student policy learning progressively through black-box level access to the teacher policy, while preserving the plasticity of the student policy.
\end{takeaway}

\section{Related Work}
Knowledge distillation~\citep{hinton2015distilling} transfers knowledge from a teacher model to a student model by minimizing the KL divergence between the two policy distributions.
Subsequently, SeqKD~\citep{kim2016sequence} transfers sequence-level information by optimizing on textual generations from the teacher model.
Symbolic KD~\citep{west2022symbolic} further extends the idea to selectively distill by designing desired prompts to elicit teacher demonstrations.
Recently, many works distill long chains of thought from stronger reasoners \citep{guo2025deepseek, ye2025limo, guha2025openthoughts} to enhance the reasoning capability of the student model.
OPD~\citep{agarwal2024policy,gu2024minillm,lu2025onpolicydistillation} transfers knowledge from the teacher model to the student model by utilizing teacher distribution over student generations, which makes use of student context during distillation to mitigate the distribution mismatch which is common in offline distillation~\citep{ross2011reduction, gu2024minillm}.
Instead of scoring student generations which fails when the capability gap is large~\citep{zhu2026many,opd2026coldstart} or at long-horizon tasks~\citep{wang2026tcod}, \method constructs supervision with student context by online teacher interventions, making it useful when student online rollouts are unreliable and when logit-level supervision is unavailable.

\section{Conclusion}
\label{sec:conclusion}
We present \method, a symbolic online distillation method that distills the teacher policy to student via online teacher intervention.
\method first collects online student prefix rollouts, introduces supervision by letting the teacher policy continue, and then calculate loss while masking out the prefix from students.
\method uses refreshed student policy at each gradient step to collect online prefix, and only uses teacher generated text as supervision source.
To boost the training efficiency, we further depoly an asynchronous implementation of \method which lets the student prefix generation and teacher continuation run in parallel.
Empirically, \method achieves 6\% to 8\% performance improvement on hard reasoning tasks and up to 22\% performance improvement on agentic benchmarks, demonstrating effective distillation performance improvement.
Further analysis also shows that \method mitigates the distillation plateauing problem in offline distillation by refreshing student context and introduces less forgetting on general benchmarks.
More broadly, our findings suggest that beyond the form of supervision, where and how supervision is placed is also an important axis of distillation design.

\section*{AI Use Statement}
We used generative AI tools to assist with polishing the writing; the authors verified all content and take full responsibility for it.

\section*{Acknowledgments}
We want to thank Alta Lab for feedback.
Dylan Zhang thanks Ilgee Hong, Vashisth Tiwari, Yapei Chang, Zhaocheng Zhu, Yuxiao Qu, Dong Yu, Haitao Mi, Michael Lee, and Xin Zhou for their discussions.

This work was supported by NSF Grant No. CHE2505932, a grant from Coefficient Giving, an Amazon AICE award, a Capital One ASKS award, and gift funding from AI2. This research also used the Delta advanced computing and data resources, which are supported by the National Science Foundation (award OAC 2005572) and the State of Illinois. Delta is a joint effort of the University of Illinois Urbana-Champaign and its National Center for Supercomputing Applications. This research used the DeltaAI advanced computing and data resource, which is supported by the National Science Foundation (award OAC 2320345) and the State of Illinois. DeltaAI is a joint effort of the University of Illinois Urbana-Champaign and its National Center for Supercomputing Applications.

\bibliographystyle{plainnat}
\bibliography{references}
\newpage
\appendix
\raggedbottom

\section{Implementation Details}
\label{app:impl}

\begin{table}[h]
    \centering
    \begin{tabular}{ll}
    \toprule
    \textbf{Hyper-parameter} & \textbf{Value} \\
    \midrule
    Training temperature                     & $1.0$ \\
    Global batch size                        & $64$ \\
    Mini batch size                           & $64$ \\
    Student rollouts per prompt             & $4$ \\
    LogProb top-$K$                 & $16$ \\
    Max response length             & $7168$ \\
    Learning rate                           & $1 \times 10^{-6}$ \\
    Train epochs                            & $1$ \\
    KL Coefficient                          & $0.0$ \\
    \bottomrule
    \end{tabular}
    \caption{Default configuration of OPD used in our reasoning tasks experiments.}
    \label{tab:opd-config}
    \end{table}

\begin{table}[h]
\centering
\begin{tabular}{ll}
\toprule
\textbf{Hyper-parameter} & \textbf{Value} \\
\midrule
Supervision signal                      & Token-level CE on teacher continuation \\
Rollout mode                            & Online, asynchronous \\
Maximum staleness                       & $3$ \\
Student rollouts per prompt             & $4$ \\
Prefix truncation point                 & $4096$ \\
Teacher continuation length             & $1024$ \\
Learning rate                           & $1 \times 10^{-5}$ \\
Train batch size                        & $64$ \\
Train epochs                            & $1$ \\
\bottomrule
\end{tabular}
\caption{Default configuration of \method used in our reasoning tasks experiments.}
\label{tab:rose-config}
\end{table}

We provide the default configuration of OPD and \method in \cref{tab:opd-config} and \cref{tab:rose-config} on reasoning tasks, respectively.
We follow the default configuration of OPD as \citet{opd2026coldstart} did.
\cref{tab:rose-config} lists the full configuration used for \method in our main experiments.
The upper block covers the roadside supervision procedure itself (student rollout, truncation, teacher continuation, and the asynchronous pipeline), and the lower block covers the optimization setup used to train the student on the resulting stitched traces.

\section{Case Studies}
\label{app:case-studies}

\subsection{Multi-turn Agentic Task}
\label{app:case-agentic}

We show a ScienceWorld~\citep{wang2022scienceworld} trajectory collected during \method training (Qwen3-1.7B student, GPT-5.4-mini teacher, 10 student turns followed by 5 teacher turns) to demonstrate online intervention works.
The student prefix and the teacher continuation are stitched into one trace.
The task asks the agent to focus on the longest-lived and then the shortest-lived animal, and the animals are placed \texttt{outside}.
For ten turns the student cycles through \texttt{go to outside}, \texttt{open outside}, and \texttt{look around}: it treats the location as the object to open and never targets the door, so the score stays at $0$.
Starting from this prefix, the teacher corrects the action to \texttt{open door to the outside} in its first turn and finishes the task four turns later.
The student therefore receives supervision on how to recover from the exact failure state it reached on its own, which a teacher-only trajectory starting from the initial observation would not contain.

\begin{promptbox}{Case Study on ScienceWorld}
\small
\begin{tabularx}{\linewidth}{@{}lX@{}}
\textbf{Task} & Find the animal with the longest life span, then the shortest life span. First, focus on the animal with the longest life span. Then, focus on the animal with the shortest life span. The animals are in the \texttt{outside} location. \\
\textbf{Setup} & Student {\color{caseStudent}\textbf{Qwen3-1.7B}} (turns 1--10) $\rightarrow$ Teacher {\color{caseTeacher}\textbf{GPT-5.4-mini}} (turns 11--15) \\
\textbf{Outcome} & Score $0.00$ at handoff $\rightarrow$ final score $1.00$ \\
\end{tabularx}

\begin{caseturn}{caseEnv}{Initial observation}{}
{\color{caseEnv}This room is called the greenhouse. In it, you see: the agent; a substance called air; a bee hive (door closed); a jug (containing nothing); a shovel; a sink, which is turned off. You also see: a door to the hallway (closed); a door to the outside (closed).}
\end{caseturn}

\begin{caseturn}{caseStudent}{Turn 1 \textperiodcentered{} Student}{score 0.00}
\caseT{I need to find the animal with the longest life span and then the shortest life span. The animals are in the `outside' location. Let me first check the `outside' location.}
\caseA{go to outside}
\caseO{The door is not open.}
\end{caseturn}

\begin{caseturn}{caseStudent}{Turn 2 \textperiodcentered{} Student}{score 0.00}
\caseT{The door to the outside is closed. I need to open it to access the animals.}
\caseA{\hlp{open outside}}
\caseO{The outside is not openable.}
\end{caseturn}

\begin{caseturn}{caseStudent}{Turn 3 \textperiodcentered{} Student}{score 0.00}
\caseT{The door to the outside is not openable. I need to check if it can be opened using another method.}
\caseA{look around}
\caseO{(same greenhouse description as the initial observation)}
\end{caseturn}

\begin{caseturn}{caseStudent}{Turns 4--10 \textperiodcentered{} Student (loop)}{score 0.00}
The student repeats turns 1--3 verbatim, with identical thoughts and observations:\par\vspace{2pt}
\begin{tabular}{@{}l@{\quad}l@{\quad}l@{}}
Turns 4--6: & \texttt{go to outside} $\rightarrow$ \texttt{\hlp{open outside}} $\rightarrow$ \texttt{look around} \\
Turns 7--9: & \texttt{go to outside} $\rightarrow$ \texttt{\hlp{open outside}} $\rightarrow$ \texttt{look around} \\
Turn 10:    & \texttt{go to outside} \quad {\color{caseEnv}(The door is not open.)} \\
\end{tabular}
\end{caseturn}

\casehandoff{Handoff: teacher continues from the student prefix (score 0.00)}

\begin{caseturn}{caseTeacher}{Turn 11 \textperiodcentered{} Teacher}{score 0.00}
\caseT{I need to open the door to the outside, not the location itself.}
\caseA{\hlg{open door to the outside}}
\caseO{The door is now open.}
\end{caseturn}

\begin{caseturn}{caseTeacher}{Turn 12 \textperiodcentered{} Teacher}{score 0.33}
\caseT{The door to the outside is now open, so I can access the outside area and find the animals.}
\caseA{go to outside}
\caseO{You move to the outside.}
\end{caseturn}

\begin{caseturn}{caseTeacher}{Turn 13 \textperiodcentered{} Teacher}{score 0.33}
\caseT{I'm in the outside area now. I need to see which animals are present and determine their life spans.}
\caseA{look around}
\caseO{This outside location is called the outside. Here you see: the agent; air; an axe; \textbf{a baby dragonfly}; a fire pit; a fountain (containing water); \textbf{a giant tortoise egg}; the ground; \textbf{a baby wolf}; wood. You also see: doors to the foundry (closed), the greenhouse (open), and the kitchen (closed).}
\end{caseturn}

\begin{caseturn}{caseTeacher}{Turn 14 \textperiodcentered{} Teacher}{score 0.83}
\caseT{The outside contains a baby dragonfly, a giant tortoise egg, and a baby wolf. The giant tortoise egg is the best candidate for the longest lifespan, so I'll focus on it first.}
\caseA{\hlg{focus on giant tortoise egg}}
\caseO{You focus on the giant tortoise egg.}
\end{caseturn}

\begin{caseturn}{caseTeacher}{Turn 15 \textperiodcentered{} Teacher}{score 1.00 \faCheck}
\caseT{Now I need to focus on the animal with the shortest life span. The baby dragonfly is the most likely shortest-lived animal here.}
\caseA{\hlg{focus on baby dragonfly}}
\caseO{You focus on the baby dragonfly.}
\end{caseturn}
\end{promptbox}

\subsection{Single-turn Reasoning on RLVE}
\label{app:case-rlve}

We also show a stitched trace on an RLVE reasoning task.
The student prefix (Qwen3-1.7B) is cut at 4096 tokens, and the teacher (Qwen3-4B-Thinking-2507) continues from the same character.
\casetrim{} marks trimmed text and everything else is verbatim.
We highlight the wrong step in the student prefix in \hlp{pink} and the correction in the teacher continuation in \hlg{green}.
This case illustrates why a short teacher continuation (1024 tokens in our main experiments) is sufficient as supervision: the teacher first completes the half-written path of the student, then corrects the error within its first few sentences.

\begin{promptbox}{Case Study on RLVE: Missed Triangle in a Cycle Count}
\small
\begin{caseturn}{caseEnv}{Problem}{gold answer: 3}
You are given an \textbf{undirected graph} with 9 vertices, labeled from \texttt{0} to \texttt{8}. The graph contains the following undirected edges: (2, 3), (2, 5), (0, 8), (1, 3), (0, 4), (0, 1), (0, 7), (3, 8), (1, 8).
Please count the number of simple cycles in the graph. A simple cycle is a cycle with at least 3 vertices, with no repeated vertices or edges. \casetrim
\end{caseturn}

\begin{caseturn}{caseStudent}{Student prefix (Qwen3-1.7B), last part before the cut}{}
\casetrim{} Wait, what about 0-4-...? No. So maybe there are only two cycles? But that seems low. Maybe I'm missing some. \casetrim\par
Vertex 1 is connected to 0,3,8. Vertex 3 is connected to 1,2,8. \casetrim{} Vertex 8 is connected to 0,3,1. \casetrim\par
\hlp{So maybe there are no other cycles. Then, the answer would be 2 cycles?} But that seems low. Let me check again. \casetrim\par
Another path: 0-1-3-8-0 (cycle), 0-1-8-0 (cycle), 0-1-3-2-5-
\end{caseturn}

\casecut

\begin{caseturn}{caseTeacher}{Teacher continuation (Qwen3-4B-Thinking-2507), first $\sim$700 characters}{answer: 3 \faCheck}
2-... no.\par
Wait, here's an idea: \hlg{what about the cycle 1-3-8-1? Let's check. 1 is connected to 3, 3 is connected to 8, 8 is connected to 1. So yes! That's a triangle.} So 1-3-8-1 is a cycle of length 3. Oh! I didn't think of that. \casetrim\par
Cycle 1: 0-1-8-0 (triangle); Cycle 2: 1-3-8-1 (triangle); Cycle 3: 0-1-3-8-0 (quadrilateral). \casetrim{} Is there a triangle 0-3-8? \casetrim{} 0 and 3 are not directly connected. \casetrim\par
{\color{caseEnv}\textit{(Within these first $\sim$700 characters, the teacher has already led the reasoning onto the correct path: the three cycles it lists are exactly the gold answer.)}}
\end{caseturn}
\end{promptbox}

\clearpage
\section{Additional Results}
\label{app:additional-results}
\subsection{Intervention Time}
\label{app:ablation-studies}
\begin{wrapfigure}{r}{0.4\textwidth}
    \centering
    \vspace{-12pt}
    \includegraphics[width=0.38\textwidth]{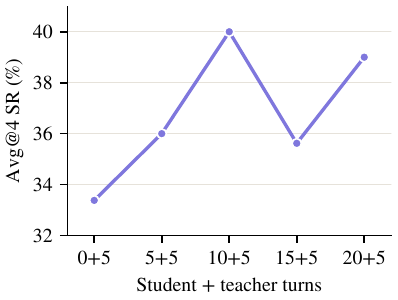}
    \caption{Avg@4 success rate of \method on ALFWorld with different numbers of student turns before the handoff.}
    \label{fig:ablation-alfworld}
    \vspace{-10pt}
\end{wrapfigure}
We perform ablation studies on \method during agentic interaction tasks on the length of student turns.
In this study, we vary the number of student turns while keeping the number of teacher turns fixed.
We use the same setup as in \cref{tab:agentic} with Qwen3-1.7B as the student and Qwen3-32B as the teacher on AlfWorld, and report the results in \cref{fig:ablation-alfworld}.
We set the number of teacher turns to 5 to study the effect of student prefix on the performance of \method.
Adding any student prefix improves the success rate, to 35.6--40.0\%, consistent with our main finding that supervision anchored at student-reachable states is more effective than supervision on off-policy teacher states.
Beyond this, performance does not increase monotonically with prefix length: 10 student turns performs best (40.0\%), while 15 and 20 turns give 35.6\% and 39.0\%.

\subsection{Generalization to Other Tasks}
\label{app:generalization-to-other-tasks}
\begin{wrapfigure}{l}{0.4\textwidth}
    \centering
    \vspace{-12pt}
    \includegraphics[width=0.38\textwidth]{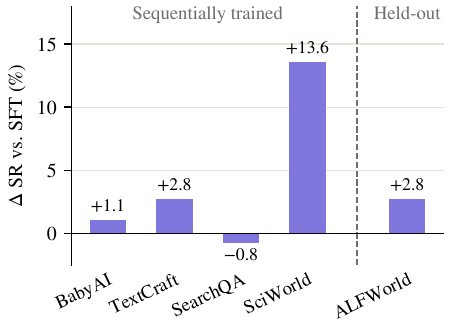}
    \caption{Success-rate gain of \method over offline distillation. \method generalizes better.}
    \label{fig:generalization}
    \vspace{-20pt}
\end{wrapfigure}
We perform another sequential training experiment to show \method can generalize to other tasks.
We start with Qwen3-1.7B as the student and GPT-5.4-mini as the teacher, and sequentially train on BabyAI, TextCraft, SearchQA and ScienceWorld.
We did not train on ALFWorld during this experiment, and use it as the held out task to evaluate the generalization ability of \method.
We report the success-rate difference between \method and offline distillation in \cref{fig:generalization}.
We observe that \method does not only persist the plasticity on sequentially trained environments, but also generalizes to other tasks.
After same amount of sequential training, \method can generalize to other tasks, with average success-rate gain over offline distillation on ALFWorld of $+2.8$.

\subsection{Additional Results on Terminal Agents}
\label{app:additional-results-terminal-agentic}
\begin{wrapfigure}{r}{0.36\textwidth}
    \centering
    \vspace{-12pt}
    \includegraphics[width=0.34\textwidth]{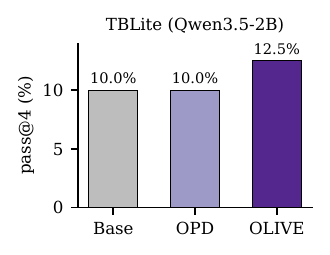}
    \caption{pass@4 on TBLite subsets with Qwen3.5-2B as the student.}
    \label{fig:tblite}
    \vspace{-10pt}
\end{wrapfigure}
We further extend the experiments on terminal agent tasks.
We select 500 training examples from TMax~\citep{ivison2026tmax}, a terminal agentic dataset that contains over 10,000 terminal agent tasks.
We use Qwen3.5-2B as the student and Qwen3.5-9B as the teacher~\citep{qwen35blog}.
For OPD, the student rolls out the first 20 turns and the teacher supervises these turns.
For \method, the student rolls out 10 turns and the teacher continues for another 10 turns, matching the 20-turn budget of OPD.
Both methods use 8 rollouts per prompt and a batch size of 16.
For evaluation, we randomly sample 50 tasks from TBLite~\citep{OpenThoughts-TBLite}, run 4 attempts per task, and report pass@4 in \cref{fig:tblite}.

As shown in \cref{fig:tblite}, OPD does not improve over the base student (10.0\% for both), while \method raises pass@4 to 12.5\%.
Terminal tasks require long horizons, and a 2B student often drifts into states from which it cannot finish the task within the first few turns.
OPD only provides token-level corrections on the student's own 20 turns, so when the student is stuck, the teacher distribution conditioned on these failing states provides little signal toward completing the task.
In contrast, the teacher continuation in \method starts from the state the student actually reaches and carries the trajectory forward, which gives the student a demonstration of how to proceed from its own intermediate states.
This is consistent with our findings on the other agentic environments in \cref{sec:agentic}.
Given the small size of the evaluation subset, we view this result as preliminary evidence that \method transfers to terminal agents.

\section{Prompts for Agentic Environments}
\label{app:agentic-prompts}

We describe the ReAct~\citep{yao2022react} interaction protocol shared by all five agentic environments in \cref{sec:agentic}, and list the instruction each environment gives to the model.
The same prompts and action parsers are used for student rollouts, teacher continuations, and evaluation.

\begin{promptlisting}{ALFWorld Instruction}
Interact with a household to solve a task. Imagine you are an intelligent agent in a household environment and your target is to perform actions to complete the task goal. At the beginning of your interactions, you will be given the detailed description of the current environment and your goal to accomplish. For each of your turn, you will be given a list of actions which you can choose one to perform in this turn. You should choose from two actions: "THOUGHT" or "ACTION". If you choose "THOUGHT", you should first think about the current condition and plan for your future actions, and then output your action in this turn. Your output must strictly follow this format:"Thought:
your thoughts.

Action:
your next action"; If you choose "ACTION", you should directly output the action in this turn. Your output must strictly follow this format:"Action:
your next action". After your each turn, the environment will give you immediate feedback based on which you plan your next few steps. if the envrionment output "Nothing happened", that means the previous action is invalid and you should try more options.
 Reminder:
1. the action must be chosen from the given available actions. Any actions except provided available actions will be regarded as illegal.
2. Think when necessary, try to act directly more in the process.
\end{promptlisting}

\begin{promptlisting}{ScienceWorld Instruction}
You are an agent for science world. Every round I will give you an observation, you have to respond an action based on the observation to finish the given task. Here are the actions you may take: [{"action": "open/close OBJ", "description": "open/close a container"}, {"action": "de/activate OBJ", "description": "activate/deactivate a device"}, {"action": "connect OBJ to OBJ", "description": "connect electrical components"}, {"action": "disconnect OBJ", "description": "disconnect electrical components"}, {"action": "use OBJ [on OBJ]", "description": "use a device/item"}, {"action": "look around", "description": "describe the current room"}, {"action": "look at OBJ", "description": "describe an object in detail"}, {"action": "look in OBJ", "description": "describe a container's contents"}, {"action": "read OBJ", "description": "read a note or book"}, {"action": "move OBJ to OBJ", "description": "move an object to a container"}, {"action": "pick up OBJ", "description": "move an object to the inventory"}, {"action": "put down OBJ", "description": "drop an inventory item"}, {"action": "pour OBJ into OBJ", "description": "pour a liquid into a container"}, {"action": "dunk OBJ into OBJ", "description": "dunk a container into a liquid"}, {"action": "mix OBJ", "description": "chemically mix a container"}, {"action": "go to LOC", "description": "move to a new location"}, {"action": "eat OBJ", "description": "eat a food"}, {"action": "flush OBJ", "description": "flush a toilet"}, {"action": "focus on OBJ", "description": "signal intent on a task object"}, {"action": "wait", "description": "take no action for 10 iterations"}, {"action": "wait1", "description": "take no action for 1 iteration"}, {"action":"examine OBJ","description":"provides a description of the objects present on or in a receptacle."}, {"action": "task", "description": "describe current task"}, {"action": "inventory", "description": "list your inventory"}]
Your response should use the following format:
Thought:
your thoughts.

Action:
your next action
\end{promptlisting}

\begin{promptlisting}{TextCraft Instruction}
You are given few useful crafting recipes to craft items in Minecraft. Crafting commands are of the format "craft [target object] using [input ingredients]".
Every round I will give you an observation, you have to respond an action based on the state and instruction. You can "get" an object (ingredients) from the inventory or the environment, look-up the game inventory by "inventory", or "craft" (target) using any of the crafting commands.
Your output must strictly follow this format:"Thought:
your thoughts.

Action:
your next action"

Reminder:
1. Always specify the quantity when using "get" and "craft" commands. - Example of get: get 1 lapis lazuli - Example1 of craft: craft 1 blue dye using 1 lapis lazuli - Example2 of craft: craft 1 golden carrot using 8 gold nugget, 1 carrot
2. When using "get" command, do not specify whether the item comes from the inventory or the environment.
3. You can use ONLY crafting commands provided, do not use your own crafting commands. However, if the crafting command uses a generic ingredient like "planks", you can use special types of the same ingredient e.g. "dark oak planks" in the command instead.
\end{promptlisting}

\begin{promptlisting}{BabyAI Instruction}
You are an exploration master that wants to finish every goal you are given. Every round I will give you an observation, and you have to respond an action and your thought based on the observation to finish the given task. You are placed in a room and you need to accomplish the given goal with actions.

You can use the following actions:

- turn right

- turn left

- move forward

- go to <obj> <id>

- pick up <obj> <id>

- go through <door> <id>: <door> must be an open door.

- toggle and go through <door> <id>: <door> can be a closed door or a locked door. If you want to open a locked door, you need to carry a key that is of the same color as the locked door.

- toggle: there is a closed or locked door right in front of you and you can toggle it.
Your response should use the following format:
Thought:
<Your Thought>

Action:
<Your Action>
\end{promptlisting}

\begin{promptlisting}{SearchQA Instruction}
You are a question-answering agent with access to a search engine over Wikipedia. Answer the question by interleaving Thought and Action steps.

Every response must use exactly this format:

Thought: <your reasoning>
Action: <one action>

There are two actions:
search[query]: search Wikipedia. The top 3 passages come back as an Observation.
answer[answer]: give your final answer, as a short phrase (an entity, name, date or number) with no explanation, e.g. answer[Beijing].

Give exactly one action per response and then stop. Do not write the Observation yourself.
\end{promptlisting}

\end{document}